\pdfoutput=1

\documentclass[11pt]{article}

\usepackage[final]{acl}

\usepackage{times}
\usepackage{latexsym}
\usepackage{multirow}
\usepackage{bm}
\usepackage{amsfonts,amssymb}
\usepackage{amsmath}
\usepackage{graphicx}
\usepackage{booktabs}

\usepackage{caption}
\usepackage{subcaption}

\usepackage[T1]{fontenc}

\usepackage[utf8]{inputenc}

\usepackage{microtype}

\title{Many-Class Text Classification with Matching}

\author{
  Yi Song$^{*}$,
  Yuxian Gu$^{*}$, 
  Minlie Huang$^{\dagger}$\\
  The CoAI group, Tsinghua University, Beijing, China \\
  Institute for Artificial Intelligence, State Key Lab of Intelligent Technology and Systems, \\
Beijing National Research Center for Information Science and Technology, \\
Department of Computer Science and Technology, Tsinghua University, Beijing, China \\
  \texttt{\{y-song21,guyx21\}@mails.tsinghua.edu.cn}\\
  \texttt{aihuang@tsinghua.edu.cn}\\
 }

\begin{document}
\maketitle
\begin{abstract}
In this work, we formulate \textbf{T}ext \textbf{C}lassification as a \textbf{M}atching problem between the text and the labels, and propose a simple yet effective framework named TCM. Compared with previous text classification approaches, TCM takes advantage of the fine-grained semantic information of the classification labels, which helps distinguish each class better when the class number is large, especially in low-resource scenarios. TCM is also easy to implement and is compatible with various large pretrained language models.
We evaluate TCM on 4 text classification datasets (each with 20+ labels) in both few-shot and full-data settings, and this model demonstrates significant improvements over other text classification paradigms. We also conduct extensive experiments with different variants of TCM and discuss the underlying factors of its success. Our method and analyses offer a new perspective on text classification.
\end{abstract}

\section{Introduction}

{\let\thefootnote\relax\footnotetext{
$^\dagger$ Corresponding author. }
\let\thefootnote\relax\footnotetext{
$^*$ indicates equal contribution. }
}



Text classification is an important task in NLP and has been widely studied a long time ago. Among text classification tasks, many-class text classification deals with the setting when the number of labels is large~\cite{multi_class}(for instance, >20), which is more challenging in practical NLP applications because the distinguish between classes is subtler with the increase of class number.

\begin{figure*}[t]
    \centering
    \includegraphics[width=\linewidth]{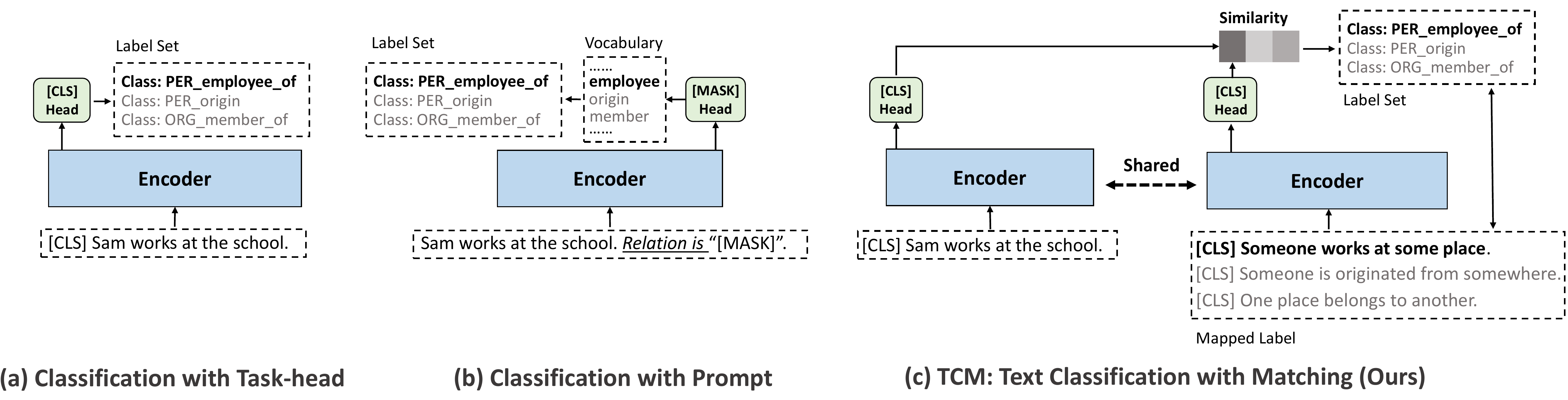}
    \caption{Two paradigms of fine-tuning pre-trained language models for text classification (Text Classification with Task-Head and Prompt) and our proposed Text Classification as Matching (TCM). In TCM, we fed the enriched labels into the encoder successively and compute the similarity between each label and the input.}
    \label{fig:framework}
\end{figure*}

Recently, thanks to the success of pre-trained language models (PLMs), fine-tuning PLMs has become a mainstream approach for various text classification tasks~\cite{plmsurvey}. The fine-tuned model inherits versatile knowledge from the pre-training corpus and shows remarkable classification performance. We illustrate two common fine-tuning approaches for text classification in Figure \ref{fig:framework}. The first one (Figure \ref{fig:framework} (a)), denoted as ``Text Classification with Task-Head'', adds a task-specific classification layer on top of PLMs and trains the classifier together with the pre-trained models~\cite{ulmfit}. The second one (Figure \ref{fig:framework} (b)) is ``Text Classification with Prompts'', which formulates text classification as a language modeling problem by inserting natural language prompts into the input. This method bridges the gap between pre-training and fine-tuning, and achieves better performance in few-shot settings~\cite{prompt_survey}.

However, classification with Task-Head usually represents classification labels using the serial numbers of the classes, which ignores their rich semantic and task-related information. Although classification with prompts maps each label to several concrete words that reflect the meaning of the corresponding class~\cite{pet, pet2, ptr}, the limited number of the mask positions restricts the use of more elaborate class information. In addition, most previous works with prompts are tested on tasks with a small number of classes (typically less than 10 classes)~\cite{lm-bff,pet,pet2,p-tuning}. In many-class text classification, the differences between the classes become more difficult and vague to distinguish.


Therefore, in this work, we propose \textbf{T}ext \textbf{C}lassification as \textbf{M}atching (TCM), a new framework for many-class text classification (Figure \ref{fig:framework} (c)). We first represent the labels with natural language sentences that explain the corresponding classes precisely. To better utilize the information of each label, we adopt a Siamese Network~\cite{siamese} to formulate text classification as a matching problem between the input texts and the class descriptions. 
As shown in Figure \ref{fig:framework}, compared to classification with Task-Head that mainly learns the class meanings from the data, TCM directly informs the model the semantic meaning of each class. And compared to classification with prompts which represents each label with several soft/discrete tokens, TCM is more flexible for introducing complex and fine-grained label information through complete sentences, which is critical for many-class text classification. TCM is also easy to implement, scalable to different language representation models of various sizes, and does not increase much inference time.

To verify the effectiveness of TCM, we conduct extensive experiments on 4 many-class classification datasets. Our findings are summarized as follows:
\begin{enumerate}
    \item Through formulating text classification as matching between the input text and class descriptions, TCM can easily incorporate complicated label information to boost model performance.
    \item Different types of class descriptions significantly affect the model performance especially under few-shot settings and for different classification tasks the proper class description content type is different.
    
    \item Though TCM is designed for text classification with massive labels, it still shows competitive performance on that with several labels.
\end{enumerate}

\section{Related Work}
\paragraph{Many-Class Text Classification}
Many-class text classification is a fundamental task in natural language processing~\cite{multi_class}. Unlike conventional classification tasks, the number of labels is large in many-class text classification, and the differences between the classes are vaguer. This requires the classifiers to capture more fine-grained class semantics. Typical many-class text classification tasks include relation extraction~\cite{fewrel}, intent detection~\cite{banking77} and large-scale emotion classification~\cite{empathetic_dialog}. To solve these tasks, some previous works adopt hierarchical methods which find the correct classes in a coarse-to-fine manner~\cite{eml_transformer}. Other works use additional data such as the task meta-data~\cite{match}. 
In contrast, our approach is simpler and does not require additional meta information.

\paragraph{Text Classification with PLMs} 
Recently, a variety of powerful PLMs such as the  GPT family~\cite{gpt,gpt2,gpt3}, BERT~\cite{bert}, RoBERTa~\cite{roberta} and T5~\cite{t5} have emerged and been adapted to downstream text classification tasks. 
Early methods add a task-specific classification head on the PLMs and train the entire model together with the additional head~\cite{ulmfit, elmo}. To bridge the gap between the pre-training and the downstream tasks, recent methods use task-related prompts to convert classification tasks to language modeling problems~\cite{pet,pet2,ptr}. To overcome the shortcomings of manually designing prompts, some works propose to automatically search the prompts~\cite{autoprompt, lm-bff}, or optimize them in a continuous space~\cite{p-tuning, prompt_tuning, ppt}. However, most of these works are only evaluated on classification tasks with a small number of labels.

\paragraph{Matching as Supervision} 
There are many works using matching as the supervision to train neural networks. In CV, models such as CLIP~\cite{clip} and ConVIRT~\cite{convirt} are trained by matching the representations of the images and their captions. These methods often yield impressive low-resource classification performance. In NLP, sentence matching is widely used in training sentence representation models~\cite{supervise_sentence_rep_nli,universal_sentence_enc,sentence_bert,simcse}. 
A commonly used architecture for encoding sentence pairs is Siamese Network~\cite{siamese}, which adopts two parameter-shared encoders(e.g., BERT) to compute the representation of each sentence. There are also works that employ the input-label matching signal for text classification~\cite{soares2019matching,wang2021entailment,liu2022pre,muller2022few}. Unlike these works, we focus on applying the matching paradigm to various classification tasks with large class numbers and comprehensively analyze the usage of class semantics, which significantly influence model performance.






\section{Text Classification as Matching}
 
\subsection{Overall Framework}
Formally, we consider many-class text classification task $\mathcal{T} = \{\mathcal{X}, \mathcal{Y}\}$ where $\mathcal{X}$ is the set of the input samples, $\mathcal{Y}$ is the set of the classes, and the $i$-th sample $x_i \in \mathcal{X}$ is annotated with a label $y_i \in \mathcal{Y}$. In TCM, we formulate any task $\mathcal{T}$ to a matching problem between the representation of $x_i$ and $y_i$. We first define a label mapping that maps each label $y_i$ to a piece of text $t_{y_i}$ consisting of concrete words. Then, we adopt a Siamese Network~\cite{siamese} as the backbone to encourage the matching between input texts and its label. In the following sections, we describe the key components of our framework in detail.




\subsection{Label Mapping}
Though there are many ways to mapping class description, we consider three label mapping approaches:
\paragraph{Class Names} Inspired by most of the prompt-based methods~\cite{pet,pet2}, we can directly map each label to its corresponding class name. However, for many-class classification tasks, simple names may not provide enough information to distinguish different classes, and it's often hard to name each class with only a few words.

\paragraph{Class Definitions} To enrich the semantic information of the labels, we try mapping each label to its class definitions. For some tasks like relation extraction, where the class definitions are naturally provided in the datasets, we directly use these definitions. For other tasks, we manually write definitions for each class.

\paragraph{Same-Class Samples} We can also assume that the meaning of each label is reflected by the samples in the corresponding class. Therefore, we also try using a randomly selected sample in each class for the label mapping.

\subsection{Matching Loss}

Then, we encourage the input text $x_i$ to match the mapped label $t_{y_i}$. We optimize the similarities between $x_i$ and $t_{y_i}$ with a cross-entropy loss:
\begin{equation} \label{loss}
\small
\mathcal{L}_{\text{m}}=-\frac{1}{N} \sum_{i = 1}^N \log \frac{\exp \left(\operatorname{sim}(x_i, t_{y_i}) / \tau\right)}{\sum_{y \in \mathcal{Y}} \exp \left(\operatorname{sim}(x_i, t_{y}) / \tau\right)},
\end{equation}
where $\tau$ is a hyper-parameter, $N$ is the total number of the training samples, and $\operatorname{sim}(x_i, t_y)$ represents the similarity between $x_i$ and $t_y$. We adopt a pre-trained encoder to get the $d$-dimension representation vectors of $x_i$ and $t_{y}$: $f_\theta(x_i)$, $f_\theta(t_{y}) \in \mathbb{R}^d$, where $\theta$ denotes the parameters of the encoder. Note that we share the parameters of the models that encode $x_i$ and $t_{y}$. Then $\operatorname{sim}(x_i, t_y) = f_{\theta}(x_i)^\top f_{\theta}(t_y)$.




\subsection{Regularization}
To help the model learn to distinguish classes with similar meanings, we also add a regularization term: 
\begin{equation}
\small
\mathcal{L}_{\text{r}}=\frac{1}{|\mathcal{Y}|} \sum_{y \in \mathcal{Y}} \max \left\{\delta, \max_{y' \in \mathcal{Y} \backslash \{y\}} \operatorname{sim}{\left(t_{y}, t_{y'}\right)}\right\},
\end{equation}
where $\delta$ is a constant threshold. By minimizing this term, the similarities between different classes are lowered, which makes them more distinguishable.

The final loss function is shown as following:
\begin{equation}
\small
\mathcal{L}=\mathcal{L}_{\text{m}}+\alpha\mathcal{L}_{\text{r}},
\label{loss function}
\end{equation}
where $\alpha$ is a hyper-parameter to balance the matching loss and the regularization.


\subsection{Inference}
During inference time, for every test sample $x$, we calculate the similarities between the input text and all the mapped labels:
\begin{equation}
    \small
    y^* = \arg \max_{y\in \mathcal{Y}} \operatorname{sim}(x, t_y).
\end{equation}

Note that after the model is trained, $f_{\theta}(t_y)$ can be pre-computed for a given task, which means the computational overhead is similar to classification with Task-Head. Compared to some work that formulates text classification as text entailment problems and concatenates the input texts and the labels~\cite{wang2021entailment}, our method is much more efficient during inference.


\begin{table*}[t]
\small
\centering
\begin{tabular}{llllllllll}
\toprule
\multirow{2}{*}{$K$}    & \multirow{2}{*}{Method} & \multicolumn{2}{c}{FewRel (80)} & \multicolumn{2}{c}{TACRED (31/41)} & \multicolumn{2}{c}{EmpatheticDialogue (32)} & \multicolumn{2}{c}{GoEmotions (28)} \\
                      &                         & BERT              & RoBERTa           & BERT             & RoBERTa          & BERT            & RoBERTa             & BERT              & RoBERTa         \\ \midrule
\multirow{3}{*}{5}    & Task-Head               & 47.8$_\text{2.5}$ & 39.3$_\text{1.5}$ & 19.5$_\text{2.3}$  & 16.1$_\text{2.4}$ & 25.3$_\text{2.1}$ & 30.3$_\text{1.4}$ & 12.7$_\text{7.8}$ & 18.0$_\text{1.3}$ \\
                      & Prompt                  & 45.1$_\text{1.2}$ & 36.7$_\text{1.0}$ & 20.7$_\text{2.1}$  & \textbf{22.4}$_\text{6.2} $& 35.5$_\text{1.4}$ & \textbf{46.8}$_\text{0.5}$ & 26.0$_\text{1.1}$ & 16.1$_\text{4.7}$  \\
                      & TCM                     & \textbf{59.4}$_\text{0.7}$ & \textbf{43.6}$_\text{0.5}$ & \textbf{22.9}$_\text{2.7}$ & 16.1$_\text{2.4}$ & \textbf{36.1}$_\text{0.9}$ & 42.9$_\text{0.9}$ & \textbf{29.2}$_\text{1.9}$ & \textbf{31.1}$_\text{1.0}$  \\ \midrule
\multirow{3}{*}{10}   & Task-Head               & 60.2$_\text{3.4}$ & 51.6$_\text{0.8}$ & 40.0$_\text{1.6}$ & 29.5$_\text{2.2}$ & 33.6$_\text{1.1}$ & 38.6$_\text{1.2}$ & 21.8$_\text{9.0}$ & 27.0$_\text{1.5}$  \\
                      & Prompt                  & 53.9$_\text{1.2}$ & 46.9$_\text{2.0}$ & 32.5$_\text{3.4}$ & \textbf{35.9}$_\text{1.7}$ & 38.1$_\text{1.0}$ & \textbf{48.4}$_\text{0.6}$ & 28.0$_\text{0.7}$ & 23.0$_\text{3.4}$ \\
                      & TCM                     & \textbf{65.6}$_\text{0.3}$ & \textbf{53.9}$_\text{1.0}$ & \textbf{42.9}$_\text{3.4}$ & 30.5$_\text{3.7}$ & \textbf{39.5}$_\text{0.6}$ & 45.4$_\text{0.3}$ & \textbf{33.6}$_\text{0.8}$ & \textbf{35.2}$_\text{1.3}$  \\ \midrule
\multirow{3}{*}{15}   & Task-Head               & 66.4$_\text{0.7}$ & 57.6$_\text{0.5}$ & 55.2$_\text{3.8}$ & 43.1$_\text{2.8}$ & 37.9$_\text{1.3}$ & 42.7$_\text{0.9}$ & 30.9$_\text{1.5}$ & 30.8$_\text{0.5}$  \\
                      & Prompt                  & 56.0$_\text{3.1}$ & 43.7$_\text{6.7}$ & 27.7$_\text{6.4}$ & 42.7$_\text{4.0}$ & 40.5$_\text{1.1}$ & \textbf{48.6}$_\text{1.0}$ & 32.4$_\text{1.4}$ & 29.7$_\text{2.3}$ \\
                      & TCM                     & \textbf{69.1}$_\text{0.9}$ & \textbf{60.1}$_\text{0.7}$ & \textbf{58.8}$_\text{2.4}$ & \textbf{44.1}$_\text{2.1}$ & \textbf{41.6}$_\text{0.4}$ & 47.4$_\text{0.7}$ & \textbf{35.4}$_\text{1.2}$ & \textbf{38.2}$_\text{0.9}$  \\ \midrule
\multirow{3}{*}{20}   & Task-Head               & 68.6$_\text{0.6}$ & 62.2$_\text{1.1}$ & 60.7$_\text{4.7}$ & 52.8$_\text{3.5}$ & 40.2$_\text{0.7}$ & 44.6$_\text{0.8}$ & 33.8$_\text{3.1}$ & 32.9$_\text{1.5}$  \\
                      & Prompt                  & 59.9$_\text{2.1}$ & 51.3$_\text{3.2}$ & 43.1$_\text{6.1}$ & 45.7$_\text{3.7}$ & 42.1$_\text{0.6}$ & \textbf{50.3}$_\text{1.1}$ & 31.7$_\text{2.1}$ & 32.8$_\text{1.6}$ \\
                      & TCM                     & \textbf{71.0}$_\text{0.6}$ & \textbf{63.4}$_\text{0.7}$ & \textbf{68.3}$_\text{2.0}$ & \textbf{56.2}$_\text{5.8}$ & \textbf{42.8}$_\text{0.3}$ & 48.2$_\text{0.8}$ & \textbf{36.9}$_\text{1.1}$ & \textbf{37.6}$_\text{1.2}$  \\ \midrule
\multirow{3}{*}{Full} & Task-Head               & 87.7              & 87.0              & 73.1              & 68.3              & 57.1         & 59.6            & 61.2             & 61.1                \\ 
                      & Prompt                  & 76.7              & 79.2              & \textbf{84.4}              & \textbf{81.8}              & 56.3         & \textbf{60.3}            & \textbf{62.3}             & \textbf{62.3}                \\
                      & TCM                     & \textbf{89.1}     & \textbf{87.6}              & 81.5              & 79.1              & \textbf{57.5}         & 59.7            & 61.8             & 62.1                  \\                 
\bottomrule
\end{tabular}
`\caption{Main experimental results. ``$K$'' represents the number of the samples corresponding to each class. The number after the dataset name means the class number(note that for TACRED, the first number menas the class number under few-shot setting and the second number means that under full-data setting). For the full-data setting ($K=\text{Full}$), we report the F1 scores on the test sets. For the few-shot setting ($K=5,10,15,20$), we report both the averaged F1 scores and the standard deviations across 5 randomly sampled training and validation sets.} \label{tab:exp_main}
\end{table*}

\section{Experiment}

\subsection{Setup}

\paragraph{Data} We conduct experiments on two datasets for relation extraction: FewRel~\cite{fewrel}, TACRED~\cite{tacred}, and two datasets for large-scale emotion classification: EmpatheticDialogue~\cite{ed}, GoEmotions~\cite{goemotions}. Each of these datasets contains more than 20 classes. Detailed data statistics can be found in the Appendix \ref{sec:appendix_datasets}. For the few-shot settings~\cite{true-few-shot}, we randomly select $K$ training samples for each class for both the training and validation sets. To reflect the uncertainty of few-shot learning, we run the experiments on 5 train/valid sets sampled with different random seeds. For the full-data setting, we randomly shuffle and split the train/valid/test sets except for TACRED on which we use the original splits.

\paragraph{Model Details} We use BERT~\cite{bert} and RoBERTa~\cite{roberta} as the pre-trained encoder. We use the class descriptions as the mapped label $t_y$ in our main experiments. A comparison of different label mappings can be found in Section \ref{sec:label_mapping}. We pass the output hidden states of the [CLS] token through an MLP layer to get the sentence representation for both the input text and the label.

\paragraph{Baselines}
We compare TCM with the two paradigms in Figure \ref{fig:framework} (a), (b): text classification with Task-Head, denoted as ``Task-Head'' and text classification with prompts, denoted as ``Prompt''. For Task-Head, we use the representation of the [CLS] token for classification. For Prompt, we mainly follow PET~\cite{pet2} to convert classification tasks to language modeling by hand-craft templates and train the model to predict the names of each class, which is widely used in current prompt-based methods. We do not use the unlabeled corpus or the ensembling tricks in PET for a fair comparison. The detail of prompt templates can be found in the Appendix \ref{sec:appendix_templates}.

\paragraph{Training Configurations}
We set batch size to 8 for few-shot setting and 32 for full-data setting.\footnote{For FewRel and TACRED datasets, we set batch size to 8 bacause of memory limitation for full-data setting.} We use the learning rate 5e-5 for $\text{BERT}_\text{BASE}$ and 2e-5 for $\text{RoBERTa}_\text{BASE}$. We adopt the AdamW optimizer and constant learning rate scheduler. 

\subsection{Main Results}
We present our main results in Table \ref{tab:exp_main}.
\paragraph{Few-shot Results}
From the rows where $K=5,10,15,20$ in Table \ref{tab:exp_main}, we have three observations. First, the Task-Head method fails to achieve satisfactory performance when the data is insufficient. Although Prompt method improves the performance on the emotion classification tasks (EmpatheticDialogue and GoEmotions), it fails on relation extraction tasks (FewRel and TACRED). The results indicate that without fine-grained information and representations of labels, the models cannot well handle the few-shot classification scenarios.

Second, TCM constantly outperforms the baselines by a large margin in all few-shot settings. The improvement is most significant when the number of the training samples is extremely low ($K=5,10$) and the performances of all methods gradually converge when $K$ increases. We also find that the performance boost of emotion classification is greater than relation extraction. We conjecture that the difference of the labels in emotion classification is vaguer than that of relation extraction and requires longer and more complex sentences to precisely represent their semantic information, where the matching paradigm of TCM shows its advantage.

Third, TCM shows a much smaller standard deviation than the baselines in most cases, which indicates that TCM is more stable across different few-shot training sets in the same task. This is probably because TCM directly informs the model the accurate semantic meanings of each label while the baselines learn the meanings of each label mainly from the datasets, resulting in a high variance of the label representations. Since few-shot learning is notorious for its instability, we conclude that TCM helps the practical use of few-shot learning by providing more reliable and robust results. 


\paragraph{Full-data Results}
We also show the results under the full-data setting ($K=\text{Full}$) in Table \ref{tab:exp_main}. We can see that although the performance of each method converges when the training samples increase, TCM still slightly outperforms the other two classification paradigms.


\subsection{Other Results}
To measure TCM performance on datasets with few class number, we evaluate it on some commonly used classification tasks. We use $\text{BERT}_\text{LARGE}$ as encoder. We use the settings and the data split from \cite{lm-bff}, except for setting learning rate to 2e-5 and batch size to 2.
Results are shown in Table \ref{tab:few-class-results}. We can see that TCM can handle the situation where the class number is few. 

\begin{table}[t]
    \small
    \centering
    \begin{tabular}{cccccc}
    \toprule
    Method           &  SST-2 & RTE  & QNLI & QQP  & MRPC     \\ 
                     &  (acc) &(acc) & (acc) & (F1)& (F1)     \\
    \midrule
    Task-Head        &  79.5  & 51.0 & 55.0 & 55.4 & 74.4      \\
    Prompt           &  85.6  & 54.2 & 54.6 & 56.5 & 66.8       \\
    TCM              &  83.1  & 53.8 & 67.9 & 60.1 & 77.5    \\
    \bottomrule
    \end{tabular}
    \caption{Few class results. 16 training samples per class. Two baseline results of SST-2 and MRPC are reported by \cite{lm-bff}, we reproduce their experiments for the other three datasets.}
    \label{tab:few-class-results}
\end{table}



\subsection{Analysis}
In this section, we further analyze the inner workings of TCM.

First, we explore the role of class description during the entire training period. On the one hand, the class description gives a reasonable initialization to the label embedding; on the other hand, it also puts some constraints when updating the label embedding because we re-encode the class description every training step. We conduct a simple experiment: just initialize the label embedding using the class description and do not use the class description again. The results are shown 
in Table \ref{tab:init}.
\begin{table}[t]
\small
\centering
\begin{tabular}{cccc}
\toprule
$K$   & Method         &   FewRel & EmpatheticDialogue \\ \midrule
5     & TCM-init       &  53.5$_\text{1.6}$      & 31.1$_\text{1.5}$   \\
5     & TCM            &  59.4$_\text{0.7}$      & 36.1$_\text{0.9}$   \\ \midrule
10    & TCM-init       &  63.7$_\text{0.8}$      & 36.3$_\text{0.6}$   \\
10    & TCM            &  65.6$_\text{0.3}$      & 39.5$_\text{0.6}$   \\ \midrule
15    & TCM-init       &  67.7$_\text{0.6}$      & 40.0$_\text{1.2}$   \\
15    & TCM            &  69.1$_\text{0.9}$      & 41.6$_\text{0.4}$   \\ \midrule
20    & TCM-init       &  70.2$_\text{0.5}$      & 41.8$_\text{0.7}$   \\
20    & TCM            &  71.0$_\text{0.6}$      & 42.8$_\text{0.3}$   \\
\bottomrule
\end{tabular}
\caption{Results of investigating class descriptions working manner. TCM-init denotes model using class descriptions just for initialization.}
\label{tab:init}
\end{table}


We can see that initializing label embeddings using class description will apparently boost model performance when the training sample is scarce, say, several per class. This shows another advantage of our matching method: the matching model can easily incorporate prior knowledge about label information.
Meanwhile, we can conclude that the major contribution of class description is during the updating rather than initializing.

\section{Ablation Study}

\begin{figure}[t]
    \centering
    \includegraphics[width=0.48\textwidth]{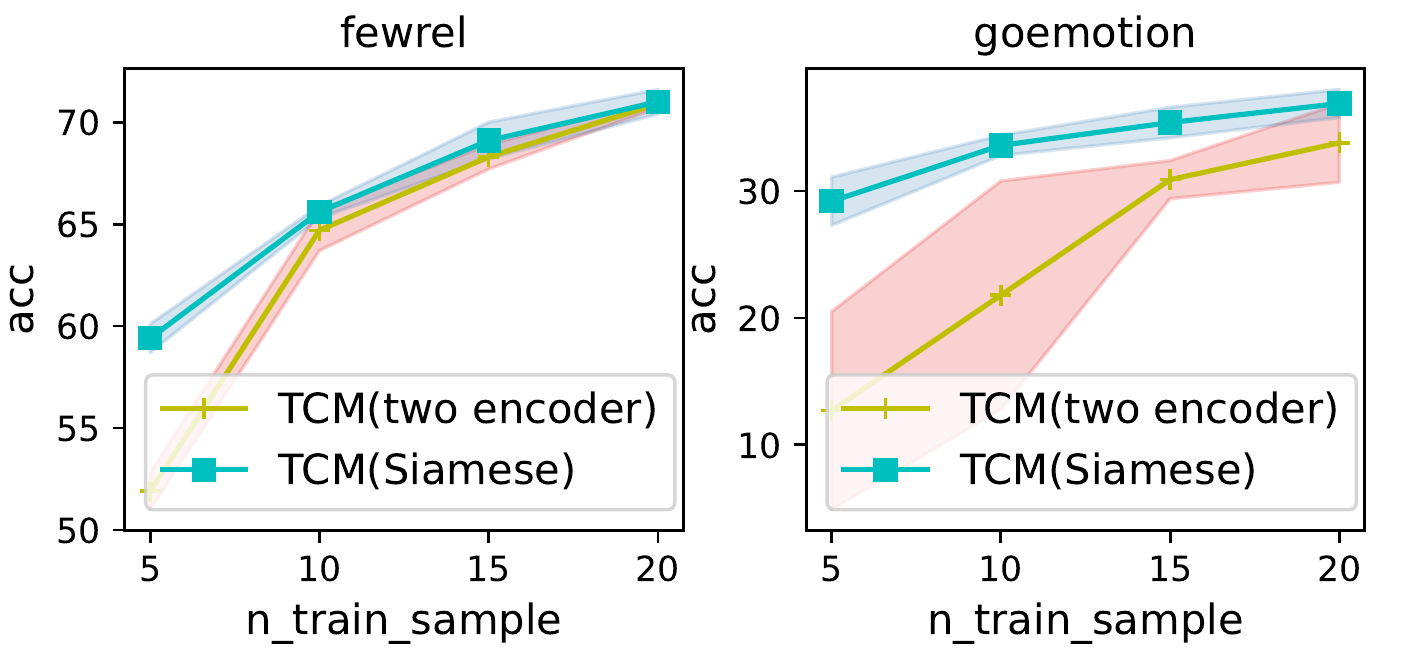}
    \caption{
    Results of two individual encoder vs. single encoder using Siamese network. The x-axis and y-axis represent the number of training sample and the accuracy score on the test set respectively. Encoder is $\text{BERT}_\text{BASE}$. }
    \label{fig:results_siamese}
\end{figure}

\paragraph{Siamese Network}
In order to verify the effectiveness of the siamese network, we use two independent encoders to encode samples and class descriptions individually. The results are shown in Figure \ref{fig:results_siamese}. We can see that the two encoder model always shows worse performance than the model using the siamese network even though the former has two times of parameters as the latter. We can also observe that the standard deviation of the two encoders is much bigger than that of the siamese network for GeEmotions dataset. This demonstrates that the siamese network is more stable under few-shot settings.

\begin{figure}[t]
    \centering
    \includegraphics[width=0.48\textwidth]{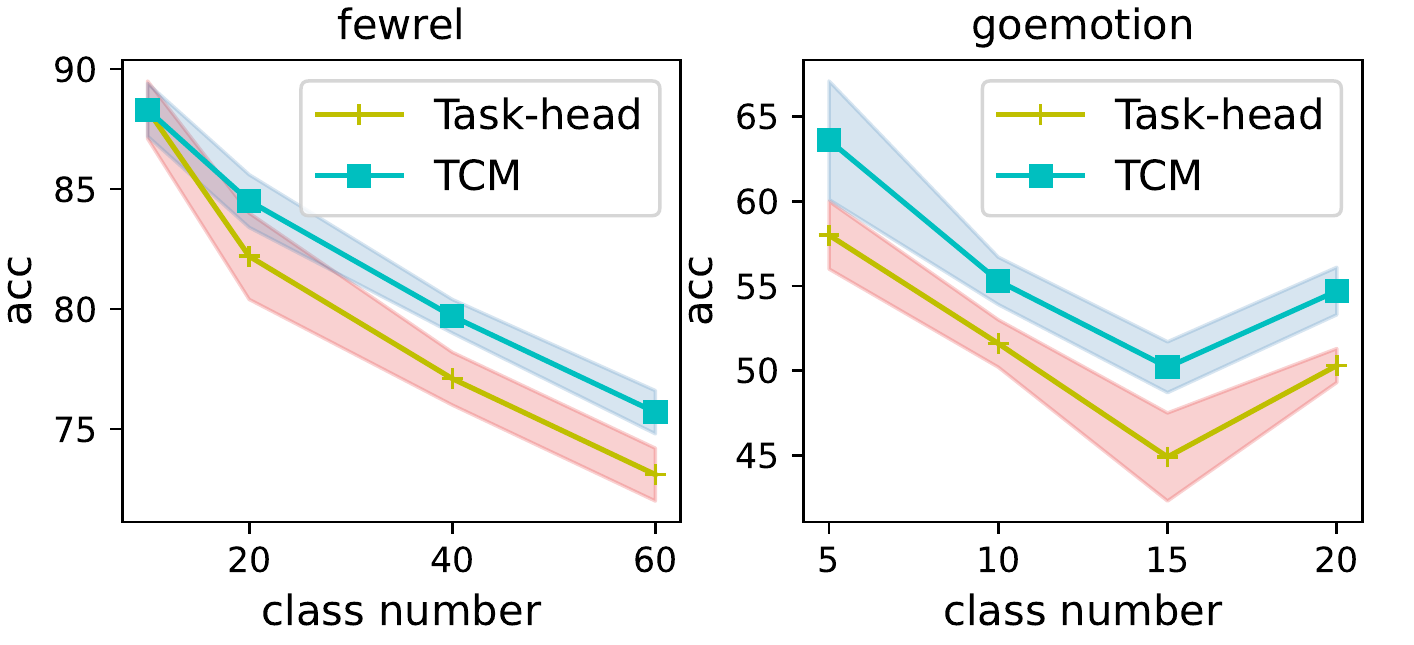}
    \caption{Results of different number of categories. The x-axis represents the number of categories. The y-axis represents the accuracy score on the test set. The experiment is under few-shot setting ($K=20$). Encoder is $\text{BERT}_\text{BASE}$. }
    \label{fig:class_number}
\end{figure}

\paragraph{Class Number}
We test the effect of the number of categories on performance, and the results are shown in Figure \ref{fig:class_number}. We use 20 samples per category and test on FewRel and EmpatheticDialogue datasets.
We can see that TCM always makes better than the Task-Head method and the gap is larger with the increase of class number. For example, when the number of classes equals 5, TCM gains a nearly equal accuracy score with the Task-Head method. However, when the number of classes increases to 20, 40, or 60, TCM can perform better than that. So we can conclude that TCM can handle both few and many categories scenarios but is more suitable for many categories.

\begin{figure}[t]
    \centering
    \includegraphics[width=0.48\textwidth]{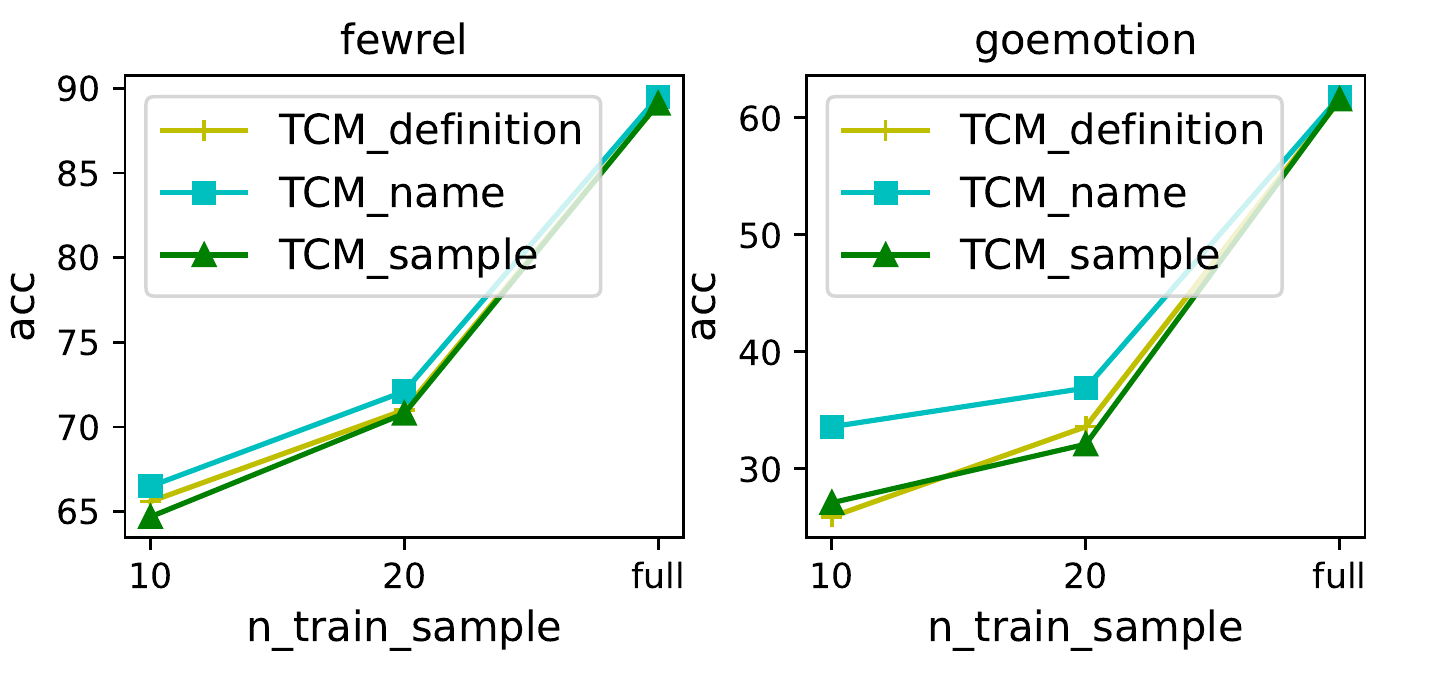}
    \caption{
    Results of different class descriptions, including definition, name, and single randomly chosen training sample. The x-axis and y-axis represent the number of training samples and the accuracy score on the test set, respectively. Encoder is $\text{BERT}_\text{BASE}$. }
    \label{fig:desc_content}
\end{figure}

\paragraph{Description Content}
\label{sec:label_mapping}
In order to explore the possible relationship between model performance and description content, we try to compare the performance of TCM with different description content, including label definition, label name, and a single training sample. Results are shown as Figure \ref{fig:desc_content}.
We can see that different class descriptions can significantly affect the model performance under few-shot setting. This experiment demonstrates that a reasonable class description like label definition or label name indeed provides some information needed to deal with the classification task. However, when the training samples are sufficient, the performance gap caused by different descriptions is faint.

\paragraph{Regularization}

\begin{figure}[t]
    \centering
    \begin{subfigure}[b]{0.48\textwidth}
        \centering
        \includegraphics[width=\textwidth]{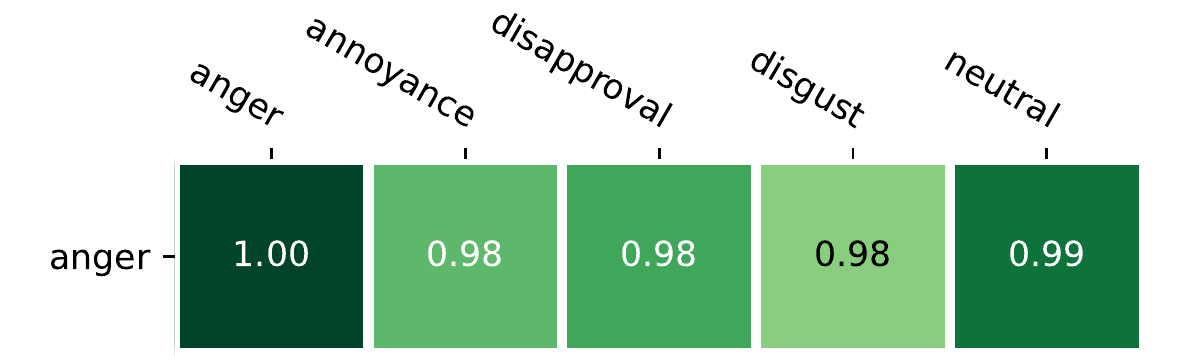}
        \caption{Few-shot setting ($K=20$).}
        \label{fig:go_ref_heat}
    \end{subfigure}
    \begin{subfigure}[b]{0.48\textwidth}
        \centering
        \includegraphics[width=\textwidth]{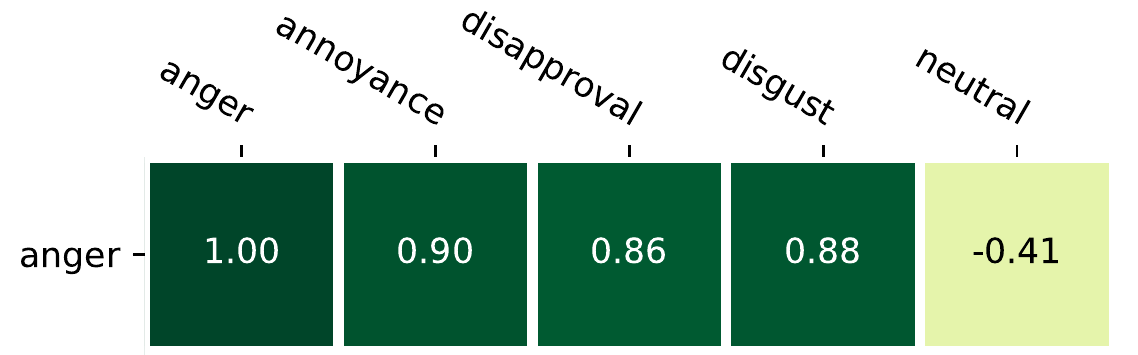}
        \caption{Full-data setting.}
        \label{fig:go_full_heat}
    \end{subfigure}
    \caption{Label embedding similarity without regularization}
    \label{fig:go_}
\end{figure}

We expect this term can effectively help the model learn to distinguish similar categories. To verify its effectiveness, we observe description embedding and confusion matrix on GoEmotions dataset. First, we disable this term during training, and the results are shown in Table \ref{tab:regularization}. 
We can see that the test accuracy is significantly dropped without this term. 
Then, we extract the label embedding from the model trained on GoEmotions dataset and $K=20$ and check their similarity. 
We can see that the label embeddings are not well distinguished. For example, seen in Figure \ref{fig:go_ref_heat}, the embedding of \textit{anger} is almost same with that of \textit{annoyance}, \textit{disapproval}, \textit{disgust} and even \textit{neutral}. 
Even if we use check this under full-data setting, it only distinguishes \textit{neutral} from the others seen in Figure \ref{fig:go_full_heat}.

\begin{table}[t]
\small
\centering
\begin{tabular}{cccc}
\toprule
$K$    &   TCM-reg              & TCM               & $\Delta$ \\ \midrule
5     &  21.1$_\text{1.2}$      & 36.1$_\text{0.9}$ & -15.0 \\ 
10    &  28.1$_\text{3.6}$      & 39.5$_\text{0.6}$ & -11.4 \\ 
15    &  29.0$_\text{1.5}$      & 41.6$_\text{0.4}$ & -12.6 \\ 
20    &  29.1$_\text{2.3}$      & 42.8$_\text{0.3}$ & -13.7 \\
Full  &  49.9               &       87.3            & -37.4 \\
\bottomrule
\end{tabular}
\caption{Results of with and without label regularization. TCM-reg denotes model trained without regularization.}
\label{tab:regularization}
\end{table}

\section{Conclusion}

This paper presents a simple yet effective framework TCM for many-class text classification. It learns the matching relation between samples and corresponding class descriptions using the siamese network and can easily incorporate prior knowledge of label information. 
Experimental results show the superior performance of TCM on different text classification tasks, especially under few-shot settings. 
We also explore how class descriptions contribute to the model and find that it gives a reasonable initialization for label embeddings and puts a constraint during parameter updating.

Ablation study shows that siamese network is essential, and using different description content will impact model performance under few-shot settings. Moreover, TCM is more suitable for large class number classification.


\section*{Acknowledgements}

\bibliography{anthology,custom}

\begin{thebibliography}{39}
\expandafter\ifx\csname natexlab\endcsname\relax\def\natexlab#1{#1}\fi

\bibitem[{Brown et~al.(2020)Brown, Mann, Ryder, Subbiah et~al.}]{gpt3}
Tom Brown, Benjamin Mann, Nick Ryder, Melanie Subbiah, et~al. 2020.
\newblock \href
  {https://papers.nips.cc/paper/2020/hash/1457c0d6bfcb4967418bfb8ac142f64a-Abstract.html}
  {Language models are few-shot learners}.
\newblock In \emph{Proceedings of NeurIPS}.

\bibitem[{Casanueva et~al.(2020)Casanueva, Temcinas, Gerz, Henderson, and
  Vulic}]{banking77}
I{\~{n}}igo Casanueva, Tadas Temcinas, Daniela Gerz, Matthew Henderson, and
  Ivan Vulic. 2020.
\newblock \href {https://arxiv.org/abs/2003.04807} {Efficient intent detection
  with dual sentence encoders}.
\newblock In \emph{Proceedings of the 2nd Workshop on NLP for ConvAI - ACL
  2020}.
\newblock Data available at
  https://github.com/PolyAI-LDN/task-specific-datasets.

\bibitem[{Cer et~al.(2018)Cer, Yang, Kong, Hua, Limtiaco, John, Constant,
  Guajardo-Cespedes, Yuan, Tar et~al.}]{universal_sentence_enc}
Daniel Cer, Yinfei Yang, Sheng-yi Kong, Nan Hua, Nicole Limtiaco, Rhomni~St
  John, Noah Constant, Mario Guajardo-Cespedes, Steve Yuan, Chris Tar, et~al.
  2018.
\newblock Universal sentence encoder.
\newblock \emph{arXiv preprint arXiv:1803.11175}.

\bibitem[{Chang et~al.(2020)Chang, Yu, Zhong, Yang, and
  Dhillon}]{eml_transformer}
Wei-Cheng Chang, Hsiang-Fu Yu, Kai Zhong, Yiming Yang, and Inderjit~S. Dhillon.
  2020.
\newblock \href {https://doi.org/10.1145/3394486.3403368} {Taming pretrained
  transformers for extreme multi-label text classification}.
\newblock In \emph{Proceedings of KDD}.

\bibitem[{Conneau et~al.(2017)Conneau, Kiela, Schwenk, Barrault, and
  Bordes}]{supervise_sentence_rep_nli}
Alexis Conneau, Douwe Kiela, Holger Schwenk, Lo{\"\i}c Barrault, and Antoine
  Bordes. 2017.
\newblock \href {https://aclanthology.org/D17-1070} {Supervised learning of
  universal sentence representations from natural language inference data}.
\newblock In \emph{Proceedings of the 2017 Conference on Empirical Methods in
  Natural Language Processing}.

\bibitem[{Demszky et~al.(2020)Demszky, Movshovitz-Attias, Ko, Cowen, Nemade,
  and Ravi}]{goemotions}
Dorottya Demszky, Dana Movshovitz-Attias, Jeongwoo Ko, Alan Cowen, Gaurav
  Nemade, and Sujith Ravi. 2020.
\newblock \href {https://aclanthology.org/2020.acl-main.372} {{G}o{E}motions: A
  dataset of fine-grained emotions}.
\newblock In \emph{Proceedings of the 58th Annual Meeting of the Association
  for Computational Linguistics}, Online.

\bibitem[{Devlin et~al.(2019)Devlin, Chang, Lee, and Toutanova}]{bert}
Jacob Devlin, Ming-Wei Chang, Kenton Lee, and Kristina Toutanova. 2019.
\newblock \href {https://aclanthology.org/N19-1423.pdf} {{BERT}: Pre-training
  of deep bidirectional transformers for language understanding}.
\newblock In \emph{Proceedings of NAACL-HLT}.

\bibitem[{Gao et~al.(2021{\natexlab{a}})Gao, Fisch, and Chen}]{lm-bff}
Tianyu Gao, Adam Fisch, and Danqi Chen. 2021{\natexlab{a}}.
\newblock \href {https://aclanthology.org/2021.acl-long.295.pdf} {Making
  pre-trained language models better few-shot learners}.
\newblock In \emph{Proceedings of ACL}.

\bibitem[{Gao et~al.(2021{\natexlab{b}})Gao, Yao, and Chen}]{simcse}
Tianyu Gao, Xingcheng Yao, and Danqi Chen. 2021{\natexlab{b}}.
\newblock Simcse: Simple contrastive learning of sentence embeddings.
\newblock \emph{arXiv preprint arXiv:2104.08821}.

\bibitem[{Gu et~al.(2022)Gu, Han, Liu, and Huang}]{ppt}
Yuxian Gu, Xu~Han, Zhiyuan Liu, and Minlie Huang. 2022.
\newblock \href {https://arxiv.org/abs/2109.04332} {{PPT}: Pre-trained prompt
  tuning for few-shor learning}.
\newblock In \emph{Proceedings of ACL}.

\bibitem[{Gupta et~al.(2014)Gupta, Bengio, and Weston}]{multi_class}
Maya~R Gupta, Samy Bengio, and Jason Weston. 2014.
\newblock Training highly multiclass classifiers.
\newblock \emph{The Journal of Machine Learning Research}, 15(1):1461--1492.

\bibitem[{Han et~al.(2021{\natexlab{a}})Han, Zhang, Ding, Gu
  et~al.}]{plmsurvey}
Xu~Han, Zhengyan Zhang, Ning Ding, Yuxian Gu, et~al. 2021{\natexlab{a}}.
\newblock \href
  {https://www.sciencedirect.com/science/article/pii/S26666510210002319}
  {Pre-trained models: Past, present and future}.
\newblock \emph{AI Open}.

\bibitem[{Han et~al.(2021{\natexlab{b}})Han, Zhao, Ding, Liu, and Sun}]{ptr}
Xu~Han, Weilin Zhao, Ning Ding, Zhiyuan Liu, and Maosong Sun.
  2021{\natexlab{b}}.
\newblock \href {https://arxiv.org/abs/2105.11259} {{PTR:} prompt tuning with
  rules for text classification}.
\newblock \emph{arXiv preprint arxiv:2105.11259}.

\bibitem[{Han et~al.(2018)Han, Zhu, Yu, Wang, Yao, Liu, and Sun}]{fewrel}
Xu~Han, Hao Zhu, Pengfei Yu, Ziyun Wang, Yuan Yao, Zhiyuan Liu, and Maosong
  Sun. 2018.
\newblock \href {https://doi.org/10.18653/v1/D18-1514} {{F}ew{R}el: A
  large-scale supervised few-shot relation classification dataset with
  state-of-the-art evaluation}.
\newblock In \emph{Proceedings of the 2018 Conference on Empirical Methods in
  Natural Language Processing}, pages 4803--4809, Brussels, Belgium.
  Association for Computational Linguistics.

\bibitem[{Howard and Ruder(2018)}]{ulmfit}
Jeremy Howard and Sebastian Ruder. 2018.
\newblock Universal language model fine-tuning for text classification.
\newblock In \emph{Proceedings of ACL}.

\bibitem[{Koch et~al.(2015)Koch, Zemel, Salakhutdinov et~al.}]{siamese}
Gregory Koch, Richard Zemel, Ruslan Salakhutdinov, et~al. 2015.
\newblock Siamese neural networks for one-shot image recognition.
\newblock In \emph{ICML deep learning workshop}, volume~2, page~0. Lille.

\bibitem[{Lester et~al.(2021)Lester, Al{-}Rfou, and Constant}]{prompt_tuning}
Brian Lester, Rami Al{-}Rfou, and Noah Constant. 2021.
\newblock \href {https://arxiv.org/abs/2104.08691} {The power of scale for
  parameter-efficient prompt tuning}.
\newblock In \emph{Proceedings of EMNLP}.

\bibitem[{Liu et~al.(2022)Liu, Lin, Han, Cao, and Sun}]{liu2022pre}
Fangchao Liu, Hongyu Lin, Xianpei Han, Boxi Cao, and Le~Sun. 2022.
\newblock Pre-training to match for unified low-shot relation extraction.
\newblock \emph{arXiv preprint arXiv:2203.12274}.

\bibitem[{Liu et~al.(2021{\natexlab{a}})Liu, Yuan, Fu, Jiang, Hayashi, and
  Neubig}]{prompt_survey}
Pengfei Liu, Weizhe Yuan, Jinlan Fu, Zhengbao Jiang, Hiroaki Hayashi, and
  Graham Neubig. 2021{\natexlab{a}}.
\newblock Pre-train, prompt, and predict: A systematic survey of prompting
  methods in natural language processing.
\newblock \emph{arXiv preprint arXiv:2107.13586}.

\bibitem[{Liu et~al.(2021{\natexlab{b}})Liu, Zheng, Du, Ding, Qian, Yang, and
  Tang}]{p-tuning}
Xiao Liu, Yanan Zheng, Zhengxiao Du, Ming Ding, Yujie Qian, Zhilin Yang, and
  Jie Tang. 2021{\natexlab{b}}.
\newblock \href {https://arxiv.org/abs/2103.10385} {{GPT} understands, too}.
\newblock \emph{arXiv preprint arXiv:2103.10385}.

\bibitem[{Liu et~al.(2019)Liu, Ott, Goyal, Du, Joshi, Chen, Levy, Lewis,
  Zettlemoyer, and Stoyanov}]{roberta}
Yinhan Liu, Myle Ott, Naman Goyal, Jingfei Du, Mandar Joshi, Danqi Chen, Omer
  Levy, Mike Lewis, Luke Zettlemoyer, and Veselin Stoyanov. 2019.
\newblock \href {https://arxiv.org/abs/1907.11692} {{RoBERTa}: A robustly
  optimized {BERT} pretraining approach}.
\newblock \emph{arXiv preprint arXiv:1907.11692}.

\bibitem[{M{\"u}ller et~al.(2022)M{\"u}ller, P{\'e}rez-Torr{\'o}, and
  Franco-Salvador}]{muller2022few}
Thomas M{\"u}ller, Guillermo P{\'e}rez-Torr{\'o}, and Marc Franco-Salvador.
  2022.
\newblock Few-shot learning with siamese networks and label tuning.
\newblock \emph{arXiv preprint arXiv:2203.14655}.

\bibitem[{Perez et~al.(2021)Perez, Kiela, and Cho}]{true-few-shot}
Ethan Perez, Douwe Kiela, and Kyunghyun Cho. 2021.
\newblock \href
  {https://proceedings.neurips.cc/paper/2021/hash/5c04925674920eb58467fb52ce4ef728-Abstract.html}
  {True few-shot learning with language models}.
\newblock In \emph{Proceedings of NeurIPS}.

\bibitem[{Peters et~al.(2018)Peters, Neumann, Iyyer, Gardner, Clark, Lee, and
  Zettlemoyer}]{elmo}
Matthew Peters, Mark Neumann, Mohit Iyyer, Matt Gardner, Christopher Clark,
  Kenton Lee, and Luke Zettlemoyer. 2018.
\newblock Deep contextualized word representations.
\newblock In \emph{Proceedings of NAACL-HLT}.

\bibitem[{Radford et~al.(2021)Radford, Kim, Hallacy, Ramesh, Goh, Agarwal,
  Sastry, Askell, Mishkin, Clark et~al.}]{clip}
Alec Radford, Jong~Wook Kim, Chris Hallacy, Aditya Ramesh, Gabriel Goh,
  Sandhini Agarwal, Girish Sastry, Amanda Askell, Pamela Mishkin, Jack Clark,
  et~al. 2021.
\newblock Learning transferable visual models from natural language
  supervision.
\newblock In \emph{International Conference on Machine Learning}, pages
  8748--8763. PMLR.

\bibitem[{Radford et~al.(2018)Radford, Narasimhan, Salimans, and
  Sutskever}]{gpt}
Alec Radford, Karthik Narasimhan, Tim Salimans, and Ilya Sutskever. 2018.
\newblock \href
  {https://www.cs.ubc.ca/~amuham01/LING530/papers/radford2018improving.pdf}
  {Improving language understanding by generative pre-training}.
\newblock \emph{OpenAI Technical report}.

\bibitem[{Radford et~al.(2019)Radford, Wu, Child, Luan, Amodei, and
  Sutskever}]{gpt2}
Alec Radford, Jeffrey Wu, Rewon Child, David Luan, Dario Amodei, and Ilya
  Sutskever. 2019.
\newblock \href {http://www.persagen.com/files/misc/radford2019language.pdf}
  {Language models are unsupervised multitask learners}.
\newblock \emph{OpenAI Technical report}.

\bibitem[{Raffel et~al.(2020)Raffel, Shazeer, Roberts, Lee, Narang, Matena,
  Zhou, Li, and Liu}]{t5}
Colin Raffel, Noam Shazeer, Adam Roberts, Katherine Lee, Sharan Narang, Michael
  Matena, Yanqi Zhou, Wei Li, and Peter~J. Liu. 2020.
\newblock \href {https://arxiv.org/abs/1910.10683} {Exploring the limits of
  transfer learning with a unified text-to-text transformer}.
\newblock \emph{JMLR}.

\bibitem[{Rashkin et~al.(2019{\natexlab{a}})Rashkin, Smith, Li, and
  Boureau}]{empathetic_dialog}
Hannah Rashkin, Eric~Michael Smith, Margaret Li, and Y-Lan Boureau.
  2019{\natexlab{a}}.
\newblock Towards empathetic open-domain conversation models: a new benchmark
  and dataset.
\newblock In \emph{ACL}.

\bibitem[{Rashkin et~al.(2019{\natexlab{b}})Rashkin, Smith, Li, and
  Boureau}]{ed}
Hannah Rashkin, Eric~Michael Smith, Margaret Li, and Y-Lan Boureau.
  2019{\natexlab{b}}.
\newblock Towards empathetic open-domain conversation models: a new benchmark
  and dataset.
\newblock In \emph{ACL}.

\bibitem[{Reimers and Gurevych(2019)}]{sentence_bert}
Nils Reimers and Iryna Gurevych. 2019.
\newblock Sentence-bert: Sentence embeddings using siamese bert-networks.
\newblock \emph{arXiv preprint arXiv:1908.10084}.

\bibitem[{Schick and Schütze(2021{\natexlab{a}})}]{pet}
Timo Schick and Hinrich Schütze. 2021{\natexlab{a}}.
\newblock \href {https://aclanthology.org/2021.eacl-main.20.pdf} {Exploiting
  cloze questions for few-shot text classification and natural language
  inference}.
\newblock In \emph{Proceedings of EACL}.

\bibitem[{Schick and Schütze(2021{\natexlab{b}})}]{pet2}
Timo Schick and Hinrich Schütze. 2021{\natexlab{b}}.
\newblock \href {https://aclanthology.org/2021.naacl-main.185.pdf} {It's not
  just size that matters: Small language models are also few-shot learners}.
\newblock In \emph{Proceedings of NAACL-HLT}.

\bibitem[{Shin et~al.(2020)Shin, Razeghi, Logan~IV, Wallace, and
  Singh}]{autoprompt}
Taylor Shin, Yasaman Razeghi, Robert~L. Logan~IV, Eric Wallace, and Sameer
  Singh. 2020.
\newblock \href {https://aclanthology.org/2020.emnlp-main.346.pdf}
  {{A}uto{P}rompt: {E}liciting {K}nowledge from {L}anguage {M}odels with
  {A}utomatically {G}enerated {P}rompts}.
\newblock In \emph{Proceedings of EMNLP}.

\bibitem[{Soares et~al.(2019)Soares, FitzGerald, Ling, and
  Kwiatkowski}]{soares2019matching}
Livio~Baldini Soares, Nicholas FitzGerald, Jeffrey Ling, and Tom Kwiatkowski.
  2019.
\newblock Matching the blanks: Distributional similarity for relation learning.
\newblock \emph{arXiv preprint arXiv:1906.03158}.

\bibitem[{Wang et~al.(2021)Wang, Fang, Khabsa, Mao, and
  Ma}]{wang2021entailment}
Sinong Wang, Han Fang, Madian Khabsa, Hanzi Mao, and Hao Ma. 2021.
\newblock Entailment as few-shot learner.
\newblock \emph{arXiv preprint arXiv:2104.14690}.

\bibitem[{Zhang et~al.(2021)Zhang, Shen, Dong, Wang, and Han}]{match}
Yu~Zhang, Zhihong Shen, Yuxiao Dong, Kuansan Wang, and Jiawei Han. 2021.
\newblock Match: Metadata-aware text classification in a large hierarchy.
\newblock In \emph{Proceedings of the Web Conference 2021}, pages 3246--3257.

\bibitem[{Zhang et~al.(2020)Zhang, Jiang, Miura, Manning, and
  Langlotz}]{convirt}
Yuhao Zhang, Hang Jiang, Yasuhide Miura, Christopher~D Manning, and Curtis~P
  Langlotz. 2020.
\newblock Contrastive learning of medical visual representations from paired
  images and text.
\newblock \emph{arXiv preprint arXiv:2010.00747}.

\bibitem[{Zhang et~al.(2017)Zhang, Zhong, Chen, Angeli, and Manning}]{tacred}
Yuhao Zhang, Victor Zhong, Danqi Chen, Gabor Angeli, and Christopher~D.
  Manning. 2017.
\newblock \href {https://aclanthology.org/D17-1004} {Position-aware attention
  and supervised data improve slot filling}.
\newblock In \emph{Proceedings of the 2017 Conference on Empirical Methods in
  Natural Language Processing}, Copenhagen, Denmark.

\end{thebibliography}
\bibliographystyle{acl_natbib}

\appendix

\section{Datasets}
\label{sec:appendix_datasets}

\paragraph{FewRel}
a few-shot relation classification dataset containing 100 relations. We rearrange its data distribution in train and valid set for experimentation, and there are only 80 available classes because test set is not accessible.

\paragraph{TACRED}
a large-scale relation extraction dataset containing 41 relation types and a "no\_relation" type. We first drop the "NA" class for all experiments and then drop 10 classes in experiments under few-shot settings because the number of samples in these classes is too small. Also, we rearrange its data in few-shot experiments.

\paragraph{EmpatheticDialogue}
a large-scale multi-turn empathetic dialogue dataset containing 32 evenly distributed emotion labels. We select the first sentence in every dialog as our sample according to its collecting principle and rearrange it. 

\paragraph{GoEmotions}
a 27 categories fine-grained emotion classification dataset contraining 12 positive, 11 negative, 4 ambiguous emotions categories and 1 "neutral". We discard all samples with multi-label and rearrange it. 


\section{Prompt Templates}
\label{sec:appendix_templates}
\paragraph{EmpatheticDialogue and GoEmotions}
for emotional classification datasets, we construct template for each sample like this: [CLS] \{sample\} [SEP] this person feels [MASK] [SEP].

\paragraph{FewRel and TACRED}
for relational classification datasets, we construct template for each sample like this: [CLS] \{sample\} [SEP]  the relation of these two entities is [MASK]*(8 for FewRel/5 for TACRED) [SEP].

\end{document}